\documentclass[letterpaper]{article} 
\usepackage{aaai2026}  
\newcommand{\Skip}[1]{}
\usepackage{times}  
\usepackage{helvet}  
\usepackage{courier}  
\usepackage[hyphens]{url}  
\usepackage{graphicx} 
\usepackage{natbib}  
\usepackage{caption} 
 \usepackage{amsfonts}
\usepackage{algorithm}
\usepackage{algorithmic}
\usepackage{array} 
 \usepackage{amsmath} 
\usepackage{newfloat}
\usepackage{listings}
\DeclareCaptionStyle{ruled}{labelfont=normalfont,labelsep=colon,strut=off} 
\floatstyle{ruled}
\newfloat{listing}{tb}{lst}{}
\floatname{listing}{Listing}
\title{Large Language Models Facilitate Vision Reflection in Image Classification}
\author{
    Guoyuan An, JaeYoon Kim, SungEui Yoon
}
\affiliations{
    School of Computing, KAIST
}

\usepackage{bibentry}

\begin{document}

\maketitle

\begin{abstract}
This paper presents several novel findings on the explainability of vision reflection on recognition in large multimodal models (LMMs). First, we show that prompting an LMM to verify the prediction of a specialized vision model can improve recognition accuracy—even on benchmarks like ImageNet—despite prior evidence that LMMs typically underperform dedicated vision encoders.
Second, we analyze the internal behavior of vision reflection and find that the vision-language connector maps visual features into explicit textual concepts, allowing the language model to reason about prediction plausibility using commonsense knowledge. We further observe that replacing a large number of vision tokens with only a few text tokens still enables LLaVA to generate similar answers, suggesting that LMMs may rely primarily on a compact set of distilled textual representations rather than raw vision features.
Third, we show that a training-free connector can enhance LMM performance in fine-grained recognition tasks, without extensive feature-alignment training.
Together, these findings offer new insights into the explainability of vision-language models and suggest that vision reflection is a promising strategy for achieving robust and interpretable visual recognition.

\end{abstract}

\section{1 Introduction.}

The capacity for \textit{reflection}, or \textit{self-correction}, is increasingly recognized as a hallmark of intelligence in large language models (LLMs), allowing them to scrutinize outputs, identify errors, and iteratively refine responses, thus enhancing performance~\cite{shinn2024reflexion, kumar2024training, huang2024self}. Recently, a few studies have begun to explore reflection in multimodal tasks~\cite{cheng2024vision, yao2024mulberry}, where visual perception serves as the primary information source. For example, accurately recognizing an obstacle on the road is essential for making reliable decisions in autonomous driving. While these works show that reflection can improve LMM performance in multimodal reasoning, they do not demonstrate improvements in core visual recognition accuracy during the reflection process. 

In contrast, numerous recent studies consistently report that LMMs typically underperform specialized vision encoders on direct classification tasks~\cite{zhang2024visually, wu2023gpt4vis, hu2024mrag, yu2025benchmarking}. For example, even advanced models like GPT-4V achieve lower accuracy (60.6\%) on ImageNet compared to dedicated vision encoders such as CLIP-L (74.8\%)~\cite{zhang2024visually}.

\begin{figure*}[th]
    \centering
    \includegraphics[width=\linewidth]{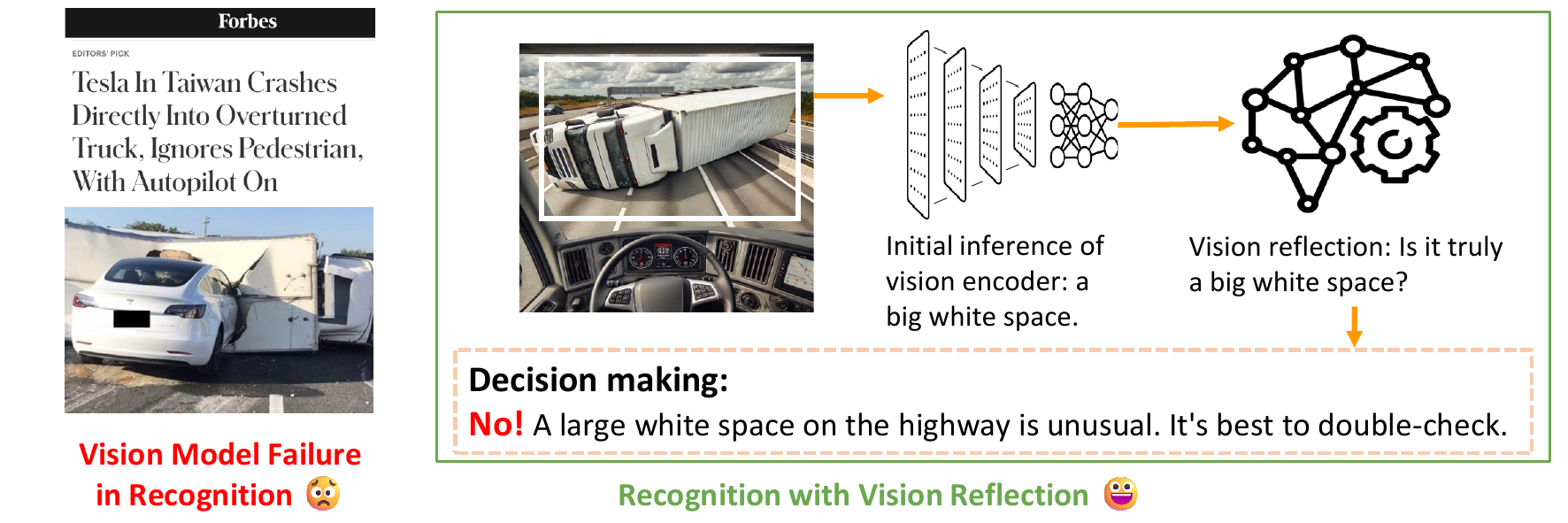}
    \caption{We propose integrating a reflection stage into the image recognition pipeline to address potential errors in the initial predictions made by the vision encoder, which may lead to undesirable outcomes. Powered by a LMM, the reflection stage allows the system to reevaluate its initial predictions and actively seek additional information, thereby mitigating risks and enhancing decision-making.} 
    \label{fig:reflection}
\end{figure*}

In this paper, we propose a two-stage classification pipeline: (1) an initial inference generated solely by a vision encoder, followed by (2) a reflective verification prompted to the LMM (e.g., ``Does the picture have a/an \{initially predicted category\}?'')~(Fig.~\ref{fig:reflection}).
We empirically demonstrate, for the first time, that this vision reflection effectively enhances accuracy on challenging benchmarks such as ImageNet. This finding is notable given prior evidence that LMMs generally perform worse than specialized vision encoders on such tasks~\cite{zhang2024visually, wu2023gpt4vis, yu2025benchmarking}. Moreover, previous efforts to improve LMM recognition through data augmentation and tuning have yielded only limited success\cite{zhang2024visually}. Thus, the ability of LMMs to improve vision encoder predictions via reflection presents a surprising and promising direction.

To understand how LMMs accomplish vision reflection despite limitations in direct classification, we investigate internal mechanisms using the LLaVA architecture as an exemplar~\cite{liu2024llava}. We design a reverse embedding layer to map vision tokens to their most relevant textual descriptors. Applying this reverse embedding layer to vision tokens reveals that the key textual terms used by the LMM to describe an image are directly encoded in these vision tokens. We replace the original image input with the extrated key textual descriptors and observe that the LMM generates similar responses. This shows that vision token features encapsulate most (though not all) of the detailed visual information in an explicit and interpretable manner, rather than relying solely on implicit representations. 

This insight sheds light on the limitations of LMMs in classification tasks: the vision-language connector module translates visual features into descriptive textual concepts. While these concepts may lack granularity for precise direct classification, they provide sufficient context for common sense reasoning about prediction plausibility. 

To further substantiate our explanation, we propose and validate a training-free approach for constructing the vision-to-language connector. This method demonstrates that by explicitly mapping image features to a vocabulary of relevant textual terms---without additional feature-alignment training---an LMM can effectively improve performance on fine-grained categories, supporting our hypothesis that LMMs transform visual features into ``text-encoding'' vision tokens through the connector.

Our primary contributions are threefold:
\begin{enumerate}
\item We provide the first evidence that LMM reflection effectively enhances image recognition accuracy on standard benchmarks like ImageNet.
\item We discover that the connector in LMMs maps vision tokens into explicit and interpretable textual concepts. Remarkably, replacing a large number of vision tokens with just a few text tokens still enables LLaVA to generate similar answers. This suggests that the LMM may primarily rely on a compact set of distilled textual concepts from the visual input. This finding also provides a novel explanation for how LMMs can perform reflective reasoning over visual predictions.
\item We demonstrate that a training-free connector can enhance LMM performance in fine-grained recognition tasks, even without extensive feature-alignment training.
\end{enumerate}


\section{2 Related Works.}

\subsection{Reflection and vision reasoning.}

\textit{Reflection}, or \textit{self-correction}, is a highly desirable capability in large language models (LLMs), referring to their ability to detect and revise their own outputs to progressively enhance response quality. Reflexion~\cite{shinn2024reflexion} demonstrates that converting scalar feedback from the environment into textual summaries allows LLMs to learn complex tasks within just a few trials. ScoRe~\cite{kumar2024training} adopts multi-turn reinforcement learning to train LLMs to correct their own mistakes; unlike Reflexion, it does not rely on ground-truth feedback during the reflection process. Huang \textit{et al.}~\cite{huang2024self} argue that LLMs contain latent ``hidden knowledge'' and show that self-improvement can be achieved through either supervised fine-tuning (SFT) or reinforcement learning with human feedback (RLHF).

Some recent studies explore visual reasoning~\cite{diwan2022winoground,thrush2022winoground,wu2023role,huang2025visionr1,wang2024measuring}, which requires step-by-step reasoning to answer multimodal queries (e.g., solving math problems with diagrams~\cite{wang2024measuring} or aligning paired images and captions~\cite{diwan2022winoground}). However, no existing work has demonstrated that large multimodal models (LMMs) can improve pure visual recognition performance through reflection. 

This paper provides the first empirical evidence that vision reflection can enhance visual recognition performance, and offers novel insights into the explainability of this process.

\subsection{LMM on vision recognition.}
\label{subsec:rel_lmm}

Large multimodal models (LMMs)\cite{achiam2023gpt,team2023gemini} combine vision and language models, demonstrating impressive capabilities in tasks such as visual question answering (VQA) and image captioning. However, to the best of our knowledge, no study has shown that integrating a language model with a trained vision model leads to improved recognition accuracy. On the contrary, recent studies\cite{zhang2024visually, wu2023gpt4vis, hu2024mrag, yu2025benchmarking} suggest that LMMs struggle with accurately identifying detailed categories, often underperforming compared to well-established vision encoders like CLIP~\cite{radford2021learning}. For instance, on the ImageNet~\cite{deng2009imagenet} benchmark, GPT-4V achieves an accuracy of only 60.6\%, while CLIP-L achieves 74.8\%, even when provided with all 1,000 possible options in the prompt.

Zhang \textit{et al.}~\cite{zhang2024visually} present comprehensive experiments showing that even strategies like prompt engineering (e.g., adding ``let's think step by step''), reducing the number of labels (e.g., limiting ImageNet’s 1,000 categories to 100 or 20 while maintaining the ground truth), and altering the inference process (e.g., having LMMs predict the probability of each candidate category rather than directly generating the class name) fail to bridge the gap between LMMs and their vision encoders. Ultimately, integrating the language model with the vision model did not yield improvements in image classification performance.

In this paper, we present, for the first time, evidence that connecting a language model with a vision model can improve recognition accuracy through vision reflection. This reflective step enhances the model's decision-making process, resulting in a meaningful improvement in classification performance on the ImageNet benchmark.



\section{3 Vision Reflection on Recognition.}
\label{sec:reflection}

\begin{figure*}[h]
    \includegraphics[width=\linewidth]{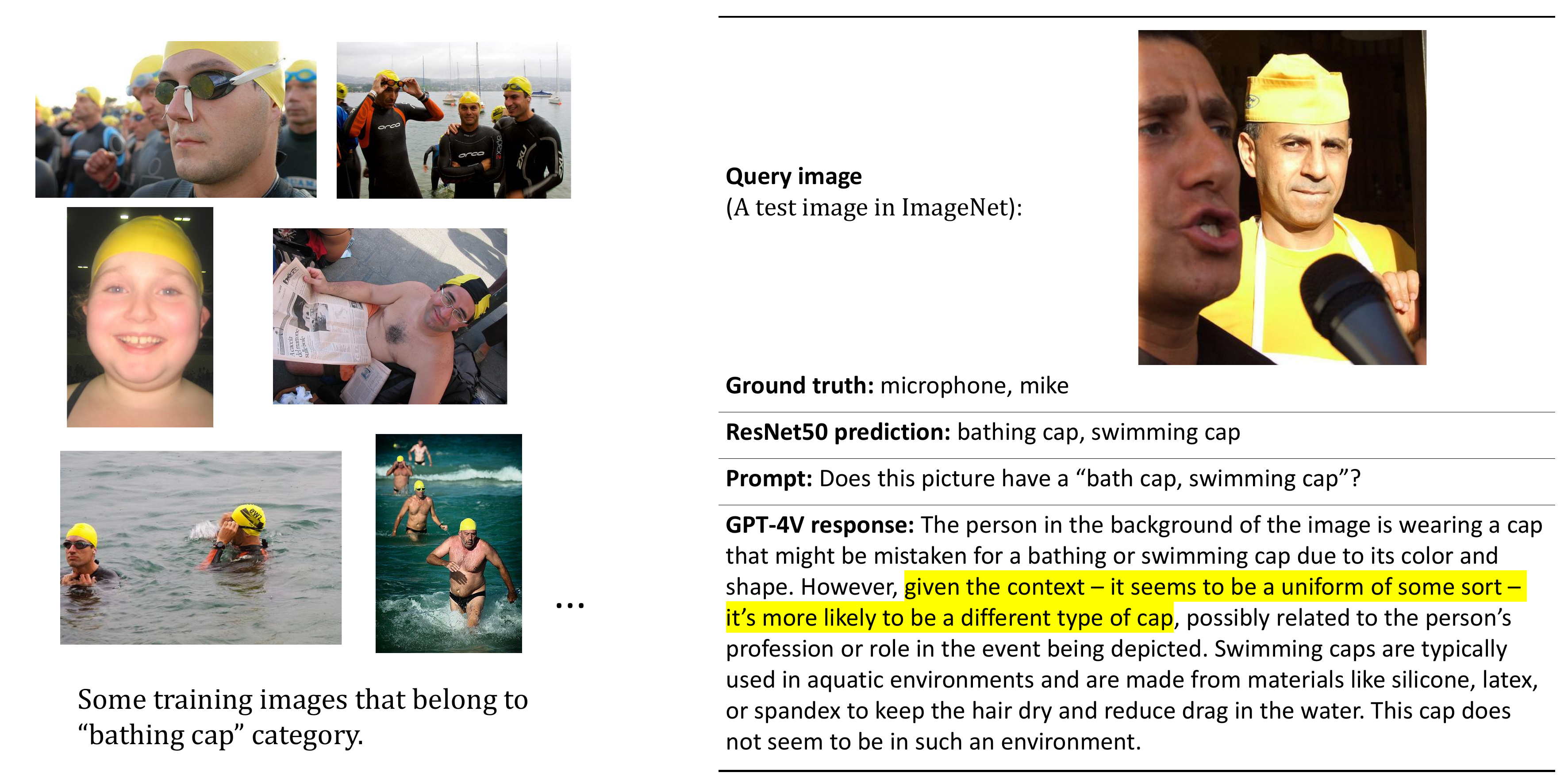}
    \vspace{-5pt}
    \caption{Qualitative example of vision reflection correcting a ResNet‑50 error.
The ResNet‑50 initially predicts “bathing cap” because the man’s cap resembles ImageNet training images of that class. The LMM reviews the prediction, articulates its rationale (sentences highlighted in \colorbox{yellow}{yellow}), and rejects the candidate. Additional qualitative results are in the Appendix.}
    \label{fig:reflection_qualitative_1}
       \vspace{-4mm}
\end{figure*}

While a few recent studies have begun to investigate reflection in multimodal tasks~\cite{cheng2024vision, yao2024mulberry}, it remains unclear whether reflection can actually enhance pure visual recognition accuracy. In fact, several works consistently find that LMMs fall short of specialized vision encoders in direct image classification tasks~\cite{zhang2024visually, wu2023gpt4vis, hu2024mrag, yu2025benchmarking}.

In this work, we divide the classification process into two stages: 1) initial inference and 2) reflection, as shown in Fig.~\ref{fig:reflection}. 
The initial inference stage relies solely on the vision encoder to predict the category, bypassing the language model. The reflection stage then revisits this prediction to verify its accuracy. This two-stage process mirrors human cognition, where an object may be misidentified at first, yet reflection enables correction. Our neuroscience insights on this mechanism are provided in Appendix.

We prompts LMMs to verify the initial prediction from the vision model by asking, ``Does the picture have a/an \{initially predicted category\}?” 
This prompt directs the LMMs to focus on a single category at a time, in contrast to prior studies~\cite{zhang2024visually,wu2023gpt4vis}, which often prompt LMMs to evaluate multiple categories simultaneously, using questions like ``Rank the \{list of categories\} from most to least likely to describe the input image” or ``What is the image? Choose one from \{list of categories\}. Let’s think step-by-step.”

Fig.~\ref{fig:reflection_qualitative_1}  illustrates an example where LMM reflection successfully identified incorrect initial inferences made by the vision model; see Appendix for more examples. LMMs appear to use common knowledge to reason and detect erroneous predictions. This common knowledge includes the familiar situations in which each category is commonly found and the definition of the object category. For instance, Fig.~\ref{fig:reflection_qualitative_1} shows a man in the background is wearing a cap. Since the cap also resembles some training images of a bathing cap from ImageNet~\cite{deng2009imagenet}, ResNet50~\cite{he2016deep}  mistakenly classifies the image as a bathing cap without contextual information. However, by reasoning about the person's profession and role in the scene, GPT-4V correctly deduces that the cap is unlikely to be a bathing cap. Furthermore, Secion~\ref{subsec:exp_imagenet} and the Appendix show quantitative and qualitative results for both the ResNet50 + GPT-4V and CLIP-ViT-L/336px~\cite{radford2021learning} + LLaVA-1.6~\cite{liu2024visual, liu2024llava} combinations. Notably, CLIP-ViT-L/336px serves as the vision encoder for LLaVA-1.6, underscoring that even with the same vision encoder, LMMs still have the potential to reflect on and identify inaccuracies in the initial inference effectively.

We observe that this reflective mechanism can also extend to video data. Examples and the binary reflection accuracy for ImageNet and Kinetics-400 subsets are reported in Section 6 and Appendix. 

The observed vision reflection ability seems to contradict previous studies showing that LMMs underperform their vision models. This leads to an important question: \textbf{If LMMs underperform their vision models, how is it possible for LMMs to improve recognition accuracy through vision reflection?}

\section{4 Why Can LMM Do Vision Reflection?}
\label{sec:why}

\begin{figure*}
    \centering
    \includegraphics[width=\linewidth]{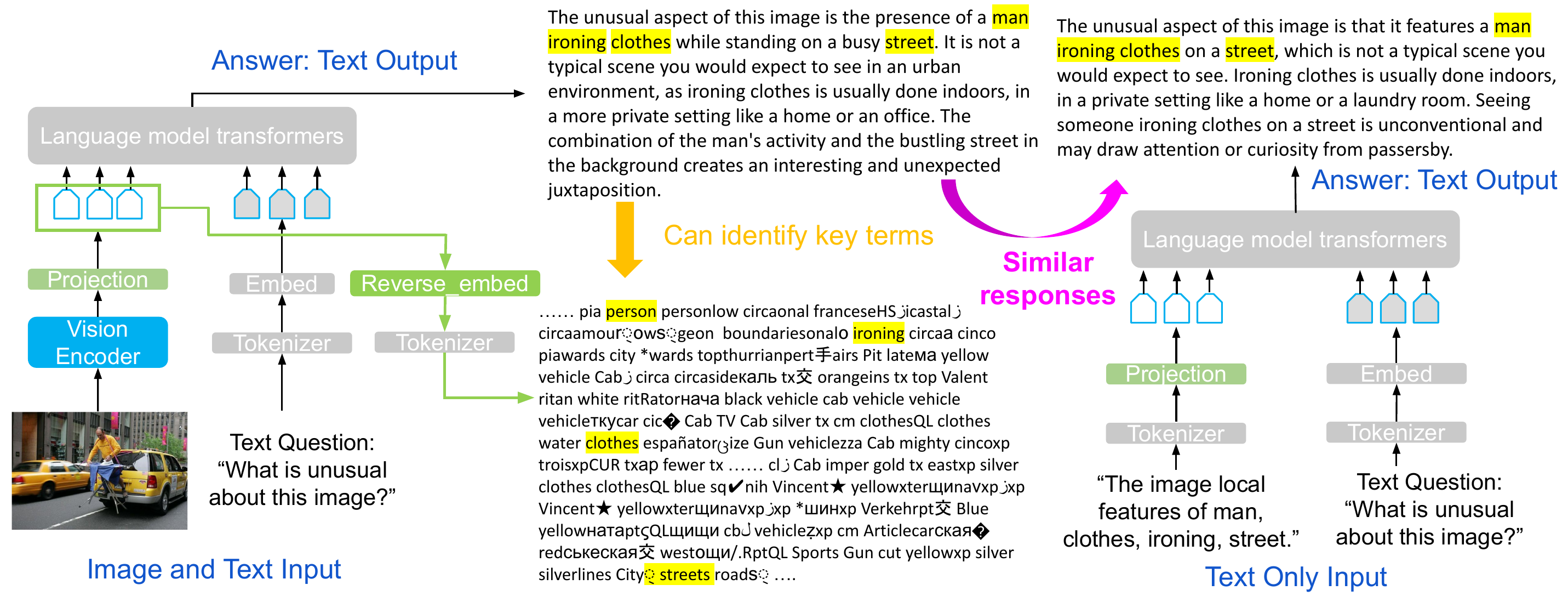}
    \caption{We implemented a reverse embedding function to convert image token features back into human-readable text. Although the resulting text appears garbled, we can still identify key terms that the language model uses to guide its responses. Interestingly, when we replace the image input with these identified text key terms, the language model generates similar answers.}
    \label{fig:connector}
       \vspace{-4mm}
\end{figure*}

It is interesting that LMMs underperform vision models in image classification but can improve the vision models' recognition accuracy through vision reflection. Note that the finding that LMMs often underperform vision models on classification tasks has already been demonstrated in previous studies through solid and extensive experiments~\cite{zhang2024visually, wu2023gpt4vis, hu2024mrag, yu2025benchmarking}, as introduced in Section~\ref{subsec:rel_lmm}. We explore the reasons behind this in this section.  

We use the LLaVA~\cite{liu2024llava} series as a representative experimental subject to explore this issue. LLaVA is one of the most popular open-source LMM architectures. It uses an MLP connector to combine the vision and language models and employs extensive feature alignment data to train connectors that transform image features from vision encoders into vision tokens interpretable by the language model~\cite{li2023blip, alayrac2022flamingo, liu2024visual, tong2024eyes, ranzinger2024radio, kar2024brave}. These vision tokens are generally understood to convey visual information implicitly within the embedding space.

We attempt to translate vision token features back into text tokens. While this might initially seem counterintuitive, we later demonstrate that this approach leads to intriguing results. The text embedding layer of the language models is typically a linear projection from tokens to an embedding space: $\mathbf{y}=\mathbf{E}(\mathbf{x})$, where $\mathbf{x} \in R^{1\times v}$ is a one-hot text token vector, $\mathbf{y} \in  R^{1\times e}$ is the corresponding text embedding, and $\mathbf{E} \in R^{v\times e}$ is a projection matrix, with $v$ as the vocabulary size and $e$ as the embedding dimension. Now given a vision token embedding $\textbf{v}$, which shares same dimension with 
$\mathbf{y}$, we design a reverse embedding function to transfer it into a text token $\mathbf{x}$ as follows:
\begin{equation}
    \mathbf{S} = \frac{\mathbf{v}}{\|\mathbf{v}\|_2} \cdot \left(\frac{\mathbf{E}}{\|\mathbf{E}\|_2}\right)^\top, 
\end{equation}

\begin{equation}
    \mathbf{x} = \text{one-hot} \left( \mathbf{S} \right), \; \mathbf{S} \in  R^{1\times v}.
\end{equation}
Here $\mathbf{S}$ records the similarity between $\mathbf{v}$ and each text token. After obtaining the token $\mathbf{x}$, we use a pre-trained text tokenizer decoder to convert them back into text. As shown in Fig.~\ref{fig:connector}, when applying the process to all vision tokens from an image, the resulting text appears garbled, as expected; however, it surprisingly reveals key terms that the language model uses to inform its responses.

Furthermore, we replace the image input with these key text terms, omitting the actual image. The key terms, formatted as ``The image local features of ...,” pass through the text tokenizer and embedding layer. (Notably, the number of tokens is much smaller than the original vision tokens.) Yet, the language model is still able to generate similar answers from when the image itself is provided as input, as shown in Fig.~\ref{fig:connector}.

While prior studies have examined the effectiveness of the linear projection layer as a connector~\cite{zhang2024visually,tong2024cambrian}, none have demonstrated that an image token can be mapped to a specific textual concept that the LMM subsequently uses to generate responses. Moreover, no previous work has shown that replacing a large number of vision tokens with just a few text tokens still allows LLaVA to produce similar answers. 

This observation helps us to better understand the LMM's vision recognition and reflection ability. LMMs struggle with direct classification because the projection layer doesn't produce enough fine-grained, relevant terms needed for classification benchmarks. However, LMMs can do vision reflection because they receive a set of descriptive terms about the image, which the language model can then use for reasoning. For example, concepts like ``bathing cap'' typically don’t co-occur with ``uniform'' or ``someone speaking at a rally'', enabling the model to reason effectively based on these conceptual associations. In other words, the LMM translates vision features into tokens ``containing textual terms'' to facilitate vision reflection. This mechanism aligns with Reflexion~\cite{shinn2024reflexion}, which transfers the original environmental signal into textual summaries to enable reflection. 
\vspace{-2mm}
\section{5 Validating Our Explanation of Vision Reflection.}
\vspace{-2mm}
\label{sec:validate}

We assess the validity of our explanation about vision reflection presented in Section~\ref{sec:why}. Our explanation suggests that LMMs perform vision reflection by converting vision features into vision tokens that encode explicit textual meaning through the connector layer. Although these vision tokens are not identical to ordinary text embeddings, a substantial portion of the information they convey is encoded in that form. Consequently, the model may also be able to process artificially constructed tokens consisting of plain descriptive terms without requiring additional feature‑alignment training. We design and conduct experiments to verify this hypothesis.

For any pre-trained vision encoder, once the image feature \( x \) (of any dimension) is obtained, we pass it through a key matrix and value matrix. The process is as follows:
\begin{equation}
\hat{x} = A(W_{\text{key}}(x)), \quad y = W_{\text{value}}(\hat{x}),
\label{eq:connector_}
\end{equation}
where \( A \) is an activation function, \( W_{\text{key}} \in \mathbb{R}^{v \times e_x} \), and \( W_{\text{value}} \in \mathbb{R}^{e_l \times v} \). Here, \( e_x \) is the dimension of the image feature \( x \), \( v \) is the vocabulary size, and \( e_l \) is the dimension of the token embedding that will be passed to the language model. 
A vision encoder’s feature is first transformed by the weight matrix \( W_{\text{key}} \) 
, resulting in an intermediate feature vector of size \( v \times 1 \) that represents the prediction scores for \( v \) categories. \( A \) is a softmax operation followed by one-hot encoding. Importantly, the vocabulary, as well as \( W_{\text{key}} \) and \( W_{\text{value}} \), can be set without additional training as following:
\begin{itemize}
    \item Vocabulary: For a pre-trained vision encoder, we create a vocabulary with terms that align with the domain in which the vision encoder excels and the specific needs of the target task. For example, this could include ImageNet categories for image classification, landmark names when the LMM serves as a tour guide for visitors, or street names when the LMM functions as a city navigator. The size of the vocabulary can vary widely, ranging from thousands to millions of terms.

    \item \( W_{\text{value}} \): For each term in the prepared vocabulary, we pass it through the tokenizer and token embedding layer of the language model. This results in \( v \) embedding vectors, each with a shape of \( 1 \times e_l \), which are then set as the parameters of \( W_{\text{value}} \).

    \item \( W_{\text{key}} \): There are three methods to obtain the weights for \( W_{\text{key}} \). For vision-language contrastive-trained encoders, such as CLIP, we use the text encoder embeddings of each term as the weights for \( W_{\text{key}} \). For vision encoders trained with classification loss, the final layer that maps hidden states to category logits can be used as the weights for \( W_{\text{key}} \). For vision encoders trained with other losses, we gather corresponding images for each category and use their vision encoder embeddings as the weights for \( W_{\text{key}} \).
\end{itemize}
\vspace{-5pt}
If our explanation in Sec 4 is correct, this training-free approach should help an LMM improve the image recognition ability on new and fine-grained categories. We show the result in Sec 6.

\section{6 Experiments.}

\subsection{Vision reflection improves recognition accuracy.}
\label{subsec:exp_imagenet}

\textbf{Dataset.} We evaluated our proposed vision reflection on the complete ImageNet validation set (50,000 images) to determine its impact on recognition accuracy, reporting results using both standard \cite{deng2009imagenet} and ReaL \cite{beyer2020we} labels.

\noindent{\textbf{Implementation.}} In our method, an initial set of top-5 candidate labels is generated by a vision model. A Large Multimodal Model (LMM) then acts as a verifier, sequentially examining each candidate from rank 1 to 5. If the LMM confirms a candidate (e.g., responding ``Yes''), that label is chosen as the final output, and the process terminates for that image. Otherwise (e.g., responding ``No'' or ``Not Sure''), the next candidate is considered. This verification loop, similar to~\cite{kumar2024training}, operates entirely without access to ground-truth information during the decision process.

To balance performance and efficiency, we apply vision reflection selectively to uncertain predictions. Specifically, we use the maximum softmax probability from the vision model’s output as a confidence score, where lower values indicate greater uncertainty. Vision reflection is then applied only to samples whose uncertainty score falls below a predefined threshold. This metric proved effective for identifying low-confidence predictions. A detailed analysis of threshold selection and its impact on overall performance is provided in the Appendix.

\newcolumntype{L}{>{\raggedright\arraybackslash}X}

\begin{table*}[ht]
  \centering
     \vspace{-4mm}
  \caption{Performances on ImageNet origin (O)~\cite{deng2009imagenet} 
           and real (R)~\cite{beyer2020we} labels. PE: Prompt Engineering~\cite{zhang2024visually}; CFG: Classifier-Free Guidance~\cite{zhang2024visually,sanchez2023stay}. *: Results obtained without an uncertainty threshold during vision reflection. }
  \setlength{\tabcolsep}{3pt}
  \renewcommand{\arraystretch}{0.97}
\hspace{-30pt}
  \begin{minipage}[t]{0.4\columnwidth}
    \centering
    \textbf{(a) LMMs}
    
    \begin{tabular}{|l|c|}
      \hline
      \textbf{Model} & \textbf{O} \\ \hline
     BLIP2-2.7B~\cite{li2023blip}
         & 25.3  \\ 
         BLIP-13B~\cite{dai2023instructblip}
         & 14.7  \\
         LLaVA-1.5-7B~\cite{liu2024visual} 
         & 22.8  \\
         LLaVA-1.6-7B ~\cite{liu2024visual} 
         & 32.3  \\
         LLaVA-1.5-13B~\cite{liu2024visual} 
         & 24.3 \\
         Claude3~\cite{team2024introducing} 
         & 53.6 \\
         GeminiPro~\cite{team2023gemini} 
         & 56.0  \\
         GPT4V~\cite{wu2023gpt4vis} 
         & 60.6 \\ \hline
    \end{tabular}
  \end{minipage}
  \hspace{110pt}
  \begin{minipage}[t]{0.39\columnwidth}
    \centering
    \textbf{(b) Prompt, Inference}

    \begin{tabular}{|l|c|}
      \hline
      \textbf{Model} & \textbf{O} \\ \hline
      LLaVA-1.5-7B 
         & 22.8  \\
     + PE 
         & 19.7 \\
         + sum tokens
         & 34.8  \\
         + CFG 
         & 47.6  \\
\hline
         BLIP2-2.7B
         & 25.3  \\ 
         + PE  
         & 27.6 \\
         + sum tokens 
         & 21.0  \\
        + CFG 
         & 38.7 \\ \hline
    \end{tabular}
  \end{minipage}
  \hspace{30pt}
  \begin{minipage}[t]{0.36\columnwidth}
    \centering
    \textbf{(c) Vision Reflection}

    \begin{tabular}{|l|c|c|}
      \hline
      \textbf{Model} & \textbf{O} & \textbf{R} \\ \hline
      ResNet‑50                          & 75.6 & 77.4 \\
      + LLaVA-1.5-7B*                  & -- & 74.4 \\
      + LLaVA-1.5-7B                  & \textbf{76.1} & \textbf{78.0} \\
      + LLaVA-1.6-34B*                 & -- & 76.9 \\ 
      + LLaVA-1.6-34B                 & \textbf{76.9} & \textbf{79.0} \\ \hline
      CLIP‑L/336p                        & 76.5 & 77.0 \\
      + LLaVA-1.6-34B*                 & -- & 75.6 \\
      + LLaVA-1.6-34B                 & \textbf{76.9} & \textbf{77.5} \\ \hline
    \end{tabular}
  \end{minipage}

  \label{tab:imagenet-all}

\end{table*}

\begin{table*}[ht]
  \centering
  \caption{Image classification accuracy on ImageNet images with a certainty score below 0.5, before and after applying reflection.}
  \vspace{-5pt}
  \setlength{\tabcolsep}{6pt}      
  \renewcommand{\arraystretch}{1.1} 
  \scalebox{0.9}{
  \begin{tabular}{|l|c|l|c|l|c|}
    \hline
    \textbf{Model} & \textbf{Accuracy} &
    \textbf{Model} & \textbf{Accuracy} &
    \textbf{Model} & \textbf{Accuracy} \\ \hline\hline

    ResNet‑50 & 33.0 & ResNet‑50 & 33.0 & CLIP‑L/336p & 38.2 \\ \hline
    +LLaVA-1.5-7B & \textbf{36.3} & + LLaVA-1.6-34B & \textbf{40.4} & + LLaVA-1.6-34B & \textbf{41.7} \\ \hline
  \end{tabular}
  }
\label{tab:reflection_uncertainty}
   \vspace{-4mm}
\end{table*}

\begin{table*}[ht]
\centering
  \caption{Performance of Reflection Ability Measured by Binary Discriminatory Metrics}
    \vspace{-10pt}
    \begin{tabular}{|c|c|c|c|c|}
    \hline
    & Accuracy & Specificity & Precision & Recall \\ \hline
         GPT4V reflection on ImageNet subset & 80.9 & 66.7 &81.3 & 89.7 \\
         Gemini reflection on Kinetics 400 subset & 93.2 & 90.5 & 91.7 & 95.7 \\
         \hline
         
    \end{tabular}
    
    \label{tab:binary}
\end{table*}

\Skip{
\begin{figure}
    \centering
    \includegraphics[width=\linewidth]{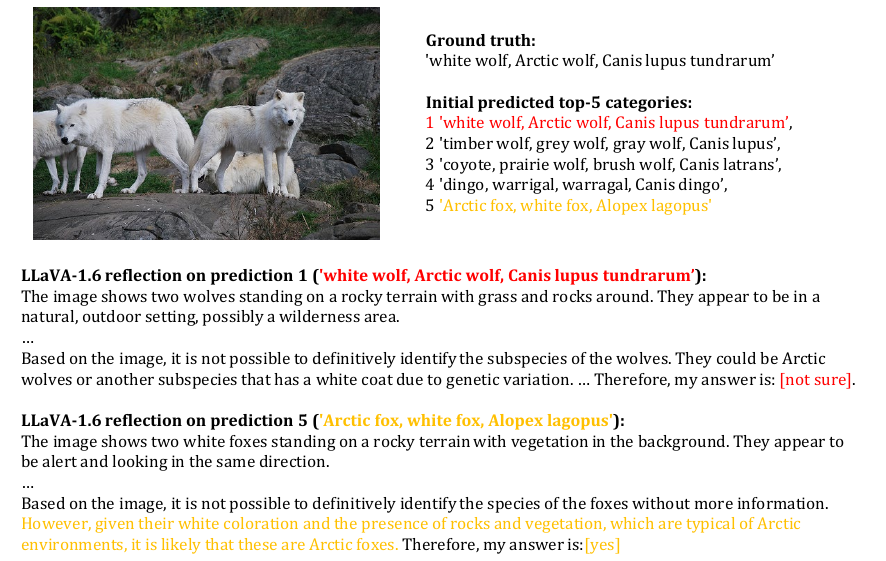}
    \caption{A failure case of LMM reflections on the initial top-5 predictions. We display only the reflection results for the 1st and 5th predictions, with the 2nd, 3rd, and 4th predictions all being ``No.'' When the LMM is unable to find a definitive reason to accept a prediction, it exhibits a degree of randomness when choosing between ``Yes,'' ``Not sure,'' or ``No.'' In this example, despite the correct initial top-1 prediction, the LMM reflection results in an incorrect reranking. 
    }
    \label{fig:wrong-case}
       \vspace{-4mm}
\end{figure}
}

\begin{figure}
    \centering
    \vspace{-3pt}
    \includegraphics[width=\linewidth]{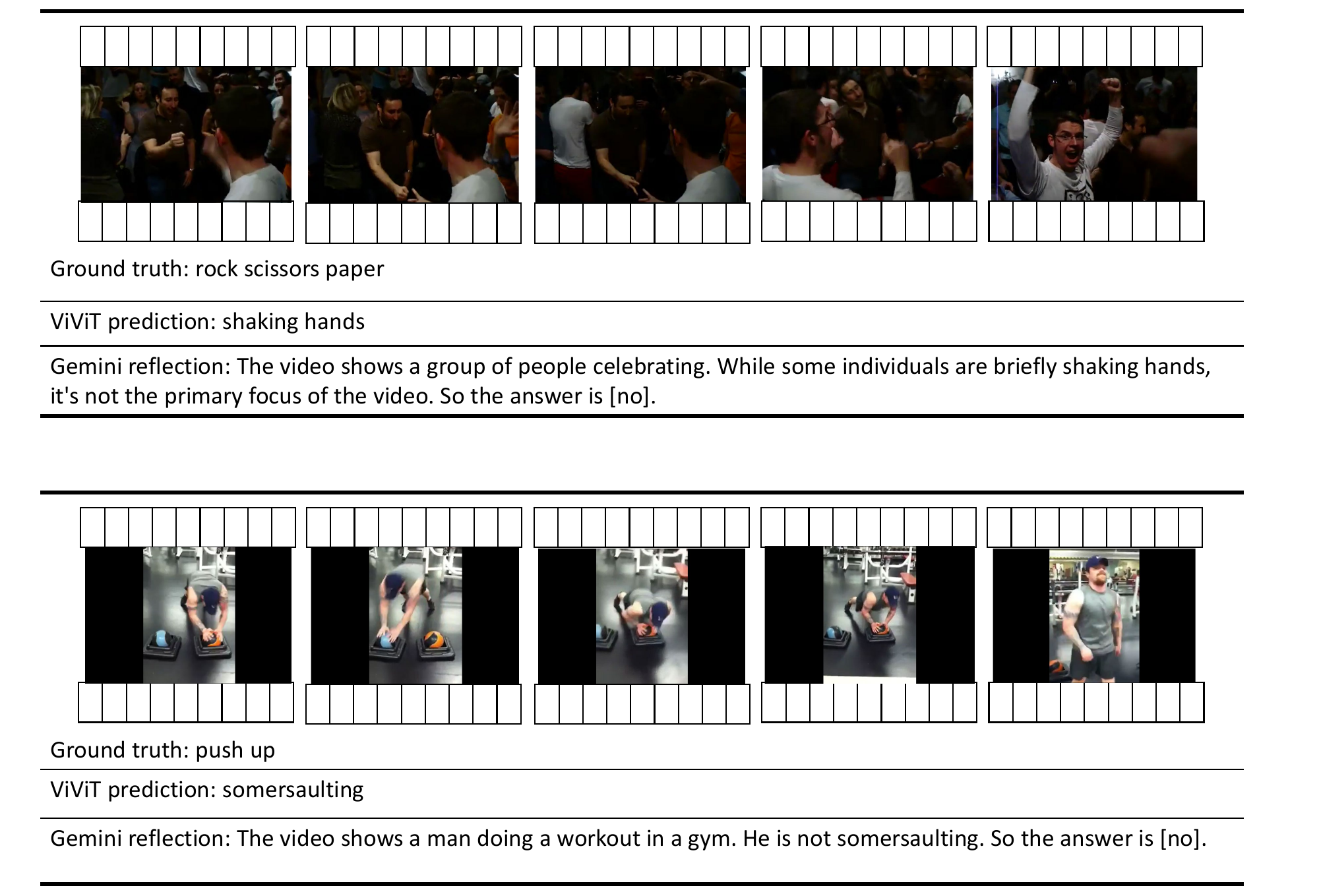}
    \vspace{-3pt}
    \caption{Two examples of Gemini performing vision reflection on video content.}
    \label{fig:reflection_video}
       \vspace{-4mm}
\end{figure}

\noindent{\textbf{Baselines.}} We benchmark against prompt engineering and inference strategies proposed in \cite{zhang2024visually}, selecting high-performing baselines. These methods typically present the LMM with both the image and a structured prompt, such as: ``Rank the \{list of categories\} from most to least likely to describe the input image.'' For detailed descriptions of each baseline, please refer to~\cite{zhang2024visually}. This comparison is appropriate for two reasons: (i) all evaluated methods ultimately rely on the LMM to make the final classification; and (ii) all operate in a zero-shot, training-free setting, relying solely on the LMM’s pre-trained capabilities.

\noindent{\textbf{Result.}} We observed that applying LMM reflection to all images directly led to a decrease in accuracy, as shown in Table~\ref{tab:imagenet-all} (entries marked with *). When the LMM cannot establish clear reasoning, it can display randomness when choosing between “Yes,” “Not sure,” or “No,” which ultimately reduces top-1 accuracy. We discuss this in Appendix. 

However, when selectively applies LMM reflection to uncertain images, performance improved significantly. As illustrated in Table~\ref{tab:imagenet-all}, with an uncertainty threshold, the accuracy of LMM reflection increased from 75.6 to 77.5, surpassing the baseline accuracy of the original vision encoder, CLIP-L/336p, which was 77.0. The impact was particularly pronounced for uncertain images selected for reflection; as shown in Table~\ref{tab:reflection_uncertainty}, LLaVA-1.6-34B reflection improved the performance of the ResNet-50 by an impressive \textbf{22.4\%} relative (from 33.0 to 40.4).

It is noteworthy that while CLIP-L/336p serves as the original vision encoder within the LLaVA-1.6-34B model, the reflection method consistently enhanced its baseline performance, achieving a 9.2\% relative increase on uncertainty images as shown in Table~\ref{tab:reflection_uncertainty}. This suggests that the observed improvements are not due to ensembling different vision encoders but are attributed to LMM reflection. Our method is also explainable; we can examine the LMM reflection content for each image, as shown in Fig.~\ref{fig:reflection_qualitative_1} and more examples in the appendix.

We evaluated its reflection capability of GPT-4V on a randomly sampled subset of 50 images with ResNet50 certainty scores below 0.99. GPT-4V improved ResNet50’s original prediction accuracy from 62\% to 78\% on this subset; note that GPT-4V's direct classification performance on ImageNet is worse than ResNet50, as shown in Table~\ref{tab:imagenet-all}. We report GPT-4V's binary discriminatory metrics (accuracy, specificity, precision, and recall) for ResNet50’s top-1 predictions in Table~\ref{tab:binary}.
We also observed the LMM can do reflection on videos, as shown in Fig.~\ref{fig:reflection_video}. We report the binary metrics from 50 random sampled videos from Kinetics-400 in Table~\ref{tab:binary}. The details are presented in appendix.

We tried various reflection prompts across different LMMs. Our findings indicate that larger models, such as GPT4V, are not influenced by prompt variations. Simple prompts like ``Is it a \{initially predicted category\}?'' or ``Is this a \{initially predicted category\}?'' effectively guide GPT-4V to do vision reflection well. In contrast, smaller models like LLaVA are sensitive to prompt wording and often generate responses that are irrelevant to the intended instructions. We discovered that repeating the category name multiple times helps LLaVA adhere to our instructions; we show examples in Appendix.

\vspace{-5pt}
\subsection{Empirical validation of the vision reflection mechanism.}
\label{subsec:exp-connector}

Sections 4 and 5 propose that LMMs achieve vision reflection by transforming visual features into ``text-encoding'' vision tokens through a connector layer. This mechanism implies that our training-free approach (detailed in Section 5) should enhance LMM image recognition on new and fine-grained categories, bypassing the need for further feature alignment training. We present the experimental validation of this approach in this subsection. 

\begin{figure}
   \vspace{-3mm}
    \centering 
    \captionof{table}{Performances on ImageWikiQA.} 
    \vspace{-4mm}
    \label{tab:imageWikiQA}
    \scalebox{1}{ 
        \begin{tabular}{|c|c|c|}
            \hline
             Model & Acc & \#param \\
             \hline
             \hline
             \multicolumn{3}{|c|}{Public LMMs} \\
             \hline
             BLIP2-2.7B~\cite{li2023blip}
             & 21.7 &2.7 billion \\ 
             IBLIP-7B~\cite{dai2023instructblip} 
             & 36.3 &7 billion \\
             IBLIP-13B~\cite{dai2023instructblip}
             & 37.5 & 13 billion\\
             LLaVA-1.6-V7B~\cite{liu2024llavanext}
             & 37.0 &7 billion\\
             LLaVA-1.5-13B~\cite{liu2023improvedllava}
             & 38.0 &13 billion\\
             LLaVA-1.6-M7B~\cite{liu2024llavanext} 
             & 40.7 & 7 billion\\
             \hline
             \hline
             \multicolumn{3}{|c|}{Proprietary LMM} \\
             \hline
             GeminiPro~\cite{team2023gemini} 
             & 49.1 & - \\
             GPT4V~\cite{wu2023gpt4vis} 
             & 61.2 &1760 billion \\
             \hline
             \hline
             \multicolumn{3}{|c|}{Ours} \\
             \hline
             Ours-LLaVA-1.6-M7B & \textbf{47.0} & 7 billion \\
            \hline
        \end{tabular}
    } 
    \vspace{-5pt}
\end{figure}

\noindent{\textbf{Dataset.}} Our primary benchmark is ImageWikiQA~\cite{zhang2024visually}, which comprises 2,000 questions, each associated with an image from ImageNet.  The questions are generated using GPT-4 based on the Wikipedia pages of ImageNet classes.
Answering questions in ImageWikiQA requires the LMM to recognize the specific types of an object instead of identifying only the general category of an image.

Additionally, we introduce two new benchmarks: GLD-VQA, derived from the Google Landmark Dataset (GLD)\cite{weyand2020google}, and Pittsburgh-VQA, from the Pittsburgh dataset\cite{arandjelovic2016netvlad}. Each new benchmark comprises 50 sampled test images. Inspired by ImageWikiQA, we developed multiple-choice questions for each image (e.g., ``Which of the following places is the location for this image?''), where the LMM must select from 10 possible answers. These benchmarks specifically assess the model's ability to recognize landmarks and urban street scenes, in contrast to ImageWikiQA’s emphasis on fine-grained species recognition.

\noindent{\textbf{Result.}} For the ImageWikiQA benchmark, we employed the 1,000 ImageNet category names as the vocabulary for our training-free connector. This connector was integrated between the vision encoder and language model of LLaVA-1.6-M7B. As detailed in Table~\ref{tab:imageWikiQA}, it enhanced the performance of LLaVANeXT-M7B by a relative 15.5\%, surpassing all experienced publicly available LMMs.

\begin{figure}
    \centering 
    \captionof{table}{Performance of LlaVA-1.6-M7B on GLD-VQA and Pitts-VQA, with and without an additional vision encoder (DELG or EigenPlaces) integrated through our training-free connector.}
        \vspace{-3pt}
        \scalebox{0.9}{
            \begin{tabular}{|c|c|c|}
                \hline
                 Model & GLD-VQA & Pitts-VQA \\
                 \hline
                 \hline
                 Origin & 38.0 & 12.0 \\
                 +DELG/EigenPlaces & \textbf{46.0} & \textbf{68.0}\\
                 \hline
            \end{tabular}
        }
        
        \label{tab:GLD-VQA}
        \vspace{-10pt}
\end{figure}
For the other two datasets, we connected DELG-ResNet101~\cite{cao2020unifying} 
and EigenPlaces-ResNet50~\cite{berton2023eigenplaces} to LLaVA-1.6-M7B via our training-free connector, using all item names from the target benchmarks as vocabularies. 
As shown in Table~\ref{tab:GLD-VQA}, the original LlaVA-1.6-M7B 
model struggles with street place recognition, achieving only 12\% accuracy, comparable to random guessing. However, integrating the new EigenPlaces vision encoder raises its performance to 68\%, demonstrating a significant improvement. For landmark domain, the original LlaVA-1.6-M7B achieves 38\% accuracy on our GLD-VQA benchmark. After adding the DELG vision encoder, this accuracy increases to 46\%.

Moreover, this integration addressed qualitative flaws. During exploratory testing with open-ended questions (distinct from the structured benchmark queries), we observed that the original LMM could produce unusual descriptions (e.g., ``human bones'' for the Radcliffe Camera). Incorporating in-domain vision encoders corrected such anomalies, as illustrated in Fig.~\ref{fig:landmark-example}.

The experimental results validate our approach: the training-free connector markedly improves LMM recognition of fine-grained categories, as expected. This enhancement lends credibility to our proposed vision reflection mechanism—specifically, its core principle that LMMs achieve this reflection by transforming visual features into ``text-encoding'' vision tokens through the connector.

\section{7 Conclusion and Future Work.}

This work pioneers the study of vision reflection for visual recognition, elucidating its mechanism and presenting the first empirical evidence of its effectiveness. We demonstrate that vision reflection can improve recognition accuracy and provide insightful explanations for this phenomenon. In particular, our findings reveal that vision tokens encapsulate explicit textual concepts, offering a deeper understanding of the explainability of vision-language models.

Vision reflection is crucial due to its potential to enhance recognition robustness and benefit applications such as autonomous driving and robotic vision. The work poses no foreseeable negative social impacts. A key limitation is the lack of extension to more multimodal tasks. Future research will focus on broadening its applications and refining the mechanisms behind visual reflection.

\Skip{
\begin{figure}[thbp] 

    \begin{minipage}[t]{0.44\linewidth} 
        \centering 
        \vspace*{-65pt} 
        \captionof{table}{Performance of LlaVA-1.6-M7B on GLD-VQA and Pitts-VQA, with and without an additional vision encoder (DELG or EigenPlaces) integrated through our training-free connector.}
        \vspace{3pt}
        \scalebox{0.8}{
            \begin{tabular}{|c|c|c|}
                \hline
                 Model & GLD-VQA & Pitts-VQA \\
                 \hline
                 \hline
                 Origin & 38.0 & 12.0 \\
                 +DELG/EigenPlaces & \textbf{46.0} & \textbf{68.0}\\
                 \hline
            \end{tabular}
        }
        
        \label{tab:GLD-VQA}
    \end{minipage}
    \hfill 
    \begin{minipage}[t]{0.54\linewidth} 
        \centering 
        \includegraphics[width=\linewidth]{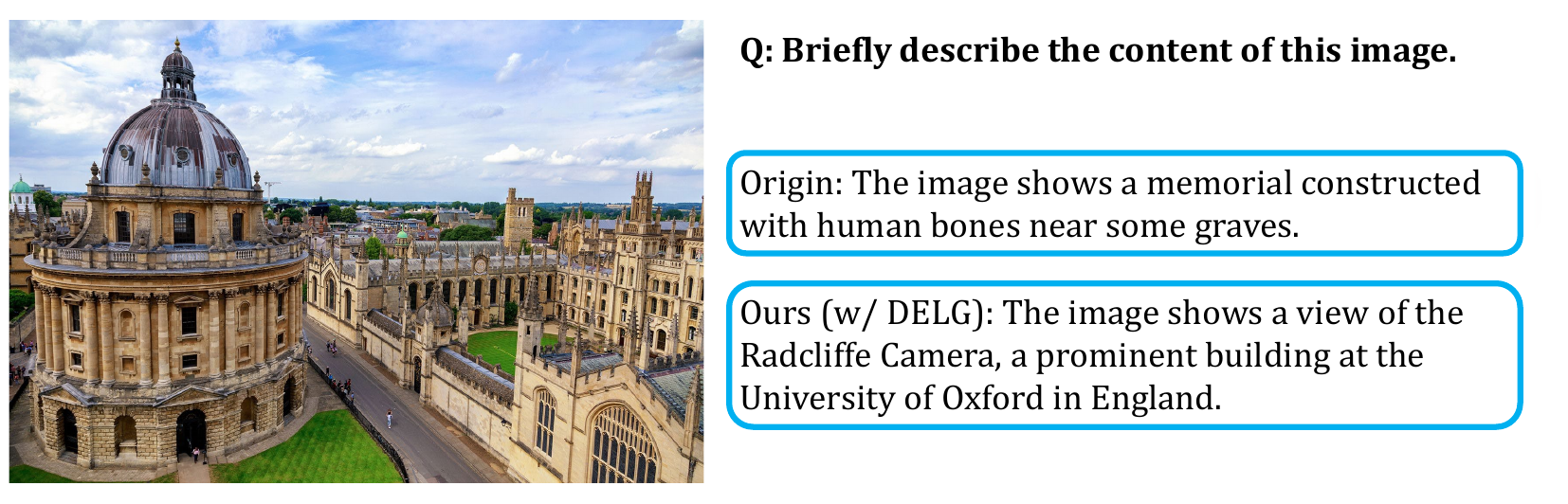} 
         \vspace{-20pt} 
        \captionof{figure}{A case demonstrating that our training-free connector enhances the VQA performance of the original LMM.}
        \label{fig:landmark-example}
    \end{minipage}
\end{figure}
}

\begin{figure}
    \centering
    \includegraphics[width=\linewidth]{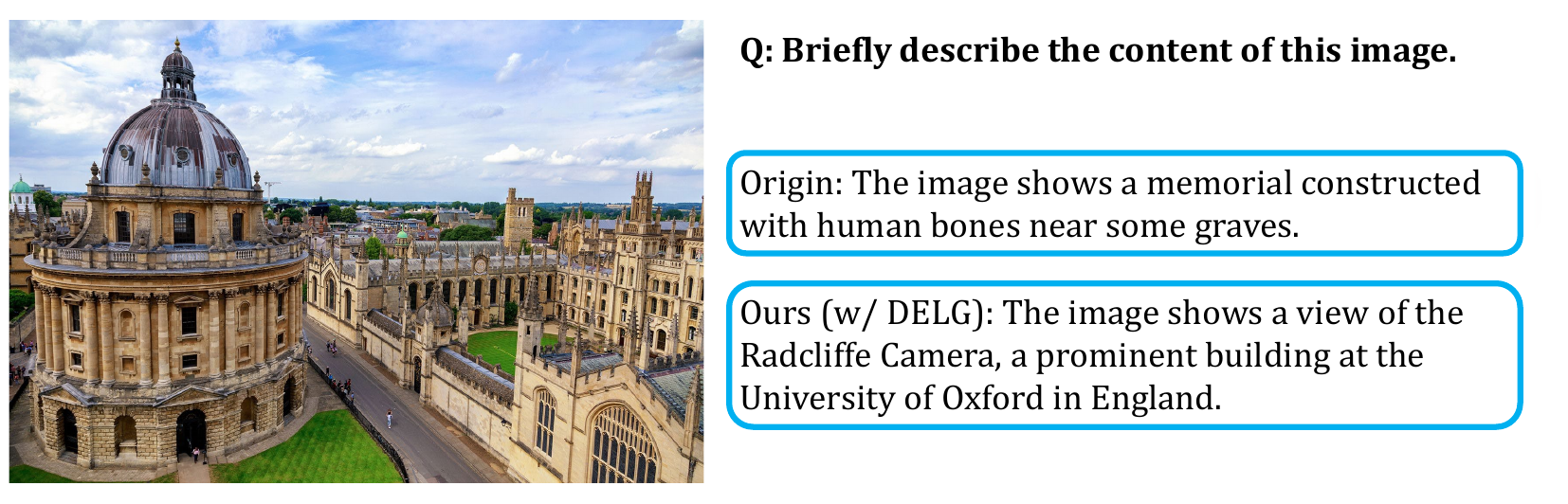}
    \caption{A case demonstrating that our training-free connector enhances the VQA performance of the original LMM on landmark domain.}
    \label{fig:landmark-example}
       \vspace{-4mm}
\end{figure}


 

\bibliography{aaai2026}

\begin{thebibliography}{37}
\providecommand{\natexlab}[1]{#1}

\bibitem[{Achiam et~al.(2023)Achiam, Adler, Agarwal, Ahmad, Akkaya, Aleman, Almeida, Altenschmidt, Altman, Anadkat et~al.}]{achiam2023gpt}
Achiam, J.; Adler, S.; Agarwal, S.; Ahmad, L.; Akkaya, I.; Aleman, F.~L.; Almeida, D.; Altenschmidt, J.; Altman, S.; Anadkat, S.; et~al. 2023.
\newblock Gpt-4 technical report.
\newblock \emph{arXiv preprint arXiv:2303.08774}.

\bibitem[{Alayrac et~al.(2022)Alayrac, Donahue, Luc, Miech, Barr, Hasson, Lenc, Mensch, Millican, Reynolds et~al.}]{alayrac2022flamingo}
Alayrac, J.-B.; Donahue, J.; Luc, P.; Miech, A.; Barr, I.; Hasson, Y.; Lenc, K.; Mensch, A.; Millican, K.; Reynolds, M.; et~al. 2022.
\newblock Flamingo: a visual language model for few-shot learning.
\newblock \emph{Advances in neural information processing systems}, 35: 23716--23736.

\bibitem[{Anthropic(2024)}]{team2024introducing}
Anthropic. 2024.
\newblock Introducing the next generation of claude.

\bibitem[{Arandjelovic et~al.(2016)Arandjelovic, Gronat, Torii, Pajdla, and Sivic}]{arandjelovic2016netvlad}
Arandjelovic, R.; Gronat, P.; Torii, A.; Pajdla, T.; and Sivic, J. 2016.
\newblock NetVLAD: CNN architecture for weakly supervised place recognition.
\newblock In \emph{Proceedings of the IEEE conference on computer vision and pattern recognition}, 5297--5307.

\bibitem[{Berton et~al.(2023)Berton, Trivigno, Caputo, and Masone}]{berton2023eigenplaces}
Berton, G.; Trivigno, G.; Caputo, B.; and Masone, C. 2023.
\newblock Eigenplaces: Training viewpoint robust models for visual place recognition.
\newblock In \emph{Proceedings of the IEEE/CVF International Conference on Computer Vision}, 11080--11090.

\bibitem[{Beyer et~al.(2020)Beyer, H{\'e}naff, Kolesnikov, Zhai, and Oord}]{beyer2020we}
Beyer, L.; H{\'e}naff, O.~J.; Kolesnikov, A.; Zhai, X.; and Oord, A. v.~d. 2020.
\newblock Are we done with imagenet?
\newblock \emph{arXiv preprint arXiv:2006.07159}.

\bibitem[{Cao, Araujo, and Sim(2020)}]{cao2020unifying}
Cao, B.; Araujo, A.; and Sim, J. 2020.
\newblock Unifying deep local and global features for image search.
\newblock In \emph{Computer Vision--ECCV 2020: 16th European Conference, Glasgow, UK, August 23--28, 2020, Proceedings, Part XX 16}, 726--743. Springer.

\bibitem[{Cheng et~al.(2024)Cheng, Li, Xu, Zhang, Zhou, and Liu}]{cheng2024vision}
Cheng, K.; Li, Y.; Xu, F.; Zhang, J.; Zhou, H.; and Liu, Y. 2024.
\newblock Vision-language models can self-improve reasoning via reflection.
\newblock \emph{arXiv preprint arXiv:2411.00855}.

\bibitem[{Dai et~al.(2023)Dai, Li, Li, Tiong, Zhao, Wang, Li, Fung, and Hoi}]{dai2023instructblip}
Dai, W.; Li, J.; Li, D.; Tiong, A. M.~H.; Zhao, J.; Wang, W.; Li, B.; Fung, P.~N.; and Hoi, S. 2023.
\newblock Instructblip: Towards general-purpose vision-language models with instruction tuning.
\newblock In \emph{NeurIPS}.

\bibitem[{Deng et~al.(2009)Deng, Dong, Socher, Li, Li, and Fei-Fei}]{deng2009imagenet}
Deng, J.; Dong, W.; Socher, R.; Li, L.-J.; Li, K.; and Fei-Fei, L. 2009.
\newblock Imagenet: A large-scale hierarchical image database.
\newblock In \emph{2009 IEEE conference on computer vision and pattern recognition}, 248--255. Ieee.

\bibitem[{Diwan et~al.(2022)Diwan, Berry, Choi, Harwath, and Mahowald}]{diwan2022winoground}
Diwan, A.; Berry, L.; Choi, E.; Harwath, D.; and Mahowald, K. 2022.
\newblock Why is Winoground Hard? Investigating Failures in Visuolinguistic Compositionality.
\newblock \emph{arXiv preprint arXiv:2211.00768}.

\bibitem[{He et~al.(2016)He, Zhang, Ren, and Sun}]{he2016deep}
He, K.; Zhang, X.; Ren, S.; and Sun, J. 2016.
\newblock Deep residual learning for image recognition.
\newblock In \emph{Proceedings of the IEEE conference on computer vision and pattern recognition}, 770--778.

\bibitem[{Hu et~al.(2024)Hu, Gu, Dou, Fayyaz, Lu, Chang, and Peng}]{hu2024mrag}
Hu, W.; Gu, J.-C.; Dou, Z.-Y.; Fayyaz, M.; Lu, P.; Chang, K.-W.; and Peng, N. 2024.
\newblock MRAG-Bench: Vision-Centric Evaluation for Retrieval-Augmented Multimodal Models.
\newblock \emph{arXiv preprint arXiv:2410.08182}.

\bibitem[{Huang et~al.(2024)Huang, Block, Foster, Rohatgi, Zhang, Simchowitz, Ash, and Krishnamurthy}]{huang2024self}
Huang, A.; Block, A.; Foster, D.~J.; Rohatgi, D.; Zhang, C.; Simchowitz, M.; Ash, J.~T.; and Krishnamurthy, A. 2024.
\newblock Self-Improvement in Language Models: The Sharpening Mechanism.
\newblock \emph{arXiv preprint arXiv:2412.01951}.

\bibitem[{Huang et~al.(2025)Huang, Jia, Zhai, Cao, Ye, Zhao, others, and Lin}]{huang2025visionr1}
Huang, W.; Jia, B.; Zhai, Z.; Cao, S.; Ye, Z.; Zhao, F.; others; and Lin, S. 2025.
\newblock Vision-R1: Incentivizing Reasoning Capability in Multimodal Large Language Models.
\newblock \emph{arXiv preprint arXiv:2503.06749}.
\newblock ArXiv:2503.06749 [cs.CL].

\bibitem[{Kar et~al.(2024)Kar, Tonioni, Poklukar, Kulshrestha, Zamir, and Tombari}]{kar2024brave}
Kar, O.~F.; Tonioni, A.; Poklukar, P.; Kulshrestha, A.; Zamir, A.; and Tombari, F. 2024.
\newblock BRAVE: Broadening the visual encoding of vision-language models.
\newblock \emph{arXiv preprint arXiv:2404.07204}.

\bibitem[{Kumar et~al.(2024)Kumar, Zhuang, Agarwal, Su, Co-Reyes, Singh, Baumli, Iqbal, Bishop, Roelofs et~al.}]{kumar2024training}
Kumar, A.; Zhuang, V.; Agarwal, R.; Su, Y.; Co-Reyes, J.~D.; Singh, A.; Baumli, K.; Iqbal, S.; Bishop, C.; Roelofs, R.; et~al. 2024.
\newblock Training language models to self-correct via reinforcement learning.
\newblock \emph{arXiv preprint arXiv:2409.12917}.

\bibitem[{Li et~al.(2023)Li, Li, Savarese, and Hoi}]{li2023blip}
Li, J.; Li, D.; Savarese, S.; and Hoi, S. 2023.
\newblock Blip-2: Bootstrapping language-image pre-training with frozen image encoders and large language models.
\newblock In \emph{International conference on machine learning}, 19730--19742. PMLR.

\bibitem[{Liu et~al.(2023)Liu, Li, Li, and Lee}]{liu2023improvedllava}
Liu, H.; Li, C.; Li, Y.; and Lee, Y.~J. 2023.
\newblock Improved Baselines with Visual Instruction Tuning.

\bibitem[{Liu et~al.(2024{\natexlab{a}})Liu, Li, Li, Li, Zhang, Shen, and Lee}]{liu2024llava}
Liu, H.; Li, C.; Li, Y.; Li, B.; Zhang, Y.; Shen, S.; and Lee, Y.~J. 2024{\natexlab{a}}.
\newblock Llava-next: Improved reasoning, ocr, and world knowledge.

\bibitem[{Liu et~al.(2024{\natexlab{b}})Liu, Li, Li, Li, Zhang, Shen, and Lee}]{liu2024llavanext}
Liu, H.; Li, C.; Li, Y.; Li, B.; Zhang, Y.; Shen, S.; and Lee, Y.~J. 2024{\natexlab{b}}.
\newblock LLaVA-NeXT: Improved reasoning, OCR, and world knowledge.

\bibitem[{Liu et~al.(2024{\natexlab{c}})Liu, Li, Wu, and Lee}]{liu2024visual}
Liu, H.; Li, C.; Wu, Q.; and Lee, Y.~J. 2024{\natexlab{c}}.
\newblock Visual instruction tuning.
\newblock \emph{Advances in neural information processing systems}, 36.

\bibitem[{Radford et~al.(2021)Radford, Kim, Hallacy, Ramesh, Goh, Agarwal, Sastry, Askell, Mishkin, Clark et~al.}]{radford2021learning}
Radford, A.; Kim, J.~W.; Hallacy, C.; Ramesh, A.; Goh, G.; Agarwal, S.; Sastry, G.; Askell, A.; Mishkin, P.; Clark, J.; et~al. 2021.
\newblock Learning transferable visual models from natural language supervision.
\newblock In \emph{International conference on machine learning}, 8748--8763. PMLR.

\bibitem[{Ranzinger et~al.(2024)Ranzinger, Heinrich, Kautz, and Molchanov}]{ranzinger2024radio}
Ranzinger, M.; Heinrich, G.; Kautz, J.; and Molchanov, P. 2024.
\newblock AM-RADIO: Agglomerative Vision Foundation Model Reduce All Domains Into One.
\newblock In \emph{Proceedings of the IEEE/CVF Conference on Computer Vision and Pattern Recognition}, 12490--12500.

\bibitem[{Sanchez et~al.(2023)Sanchez, Fan, Spangher, Levi, Ammanamanchi, and Biderman}]{sanchez2023stay}
Sanchez, G.; Fan, H.; Spangher, A.; Levi, E.; Ammanamanchi, P.~S.; and Biderman, S. 2023.
\newblock Stay on topic with classifier-free guidance.
\newblock \emph{arXiv preprint arXiv:2306.17806}.

\bibitem[{Shinn et~al.(2024)Shinn, Cassano, Gopinath, Narasimhan, and Yao}]{shinn2024reflexion}
Shinn, N.; Cassano, F.; Gopinath, A.; Narasimhan, K.; and Yao, S. 2024.
\newblock Reflexion: Language agents with verbal reinforcement learning.
\newblock \emph{Advances in Neural Information Processing Systems}, 36.

\bibitem[{Team et~al.(2023)Team, Anil, Borgeaud, Wu, Alayrac, Yu, Soricut, Schalkwyk, Dai, Hauth et~al.}]{team2023gemini}
Team, G.; Anil, R.; Borgeaud, S.; Wu, Y.; Alayrac, J.-B.; Yu, J.; Soricut, R.; Schalkwyk, J.; Dai, A.~M.; Hauth, A.; et~al. 2023.
\newblock Gemini: a family of highly capable multimodal models.
\newblock \emph{arXiv preprint arXiv:2312.11805}.

\bibitem[{Thrush et~al.(2022)Thrush, Jiang, Bartolo, Singh, Williams, Kiela, and Ross}]{thrush2022winoground}
Thrush, T.; Jiang, R.; Bartolo, M.; Singh, A.; Williams, A.; Kiela, D.; and Ross, C. 2022.
\newblock Winoground: Probing vision and language models for visio-linguistic compositionality.
\newblock In \emph{Proceedings of the IEEE/CVF Conference on Computer Vision and Pattern Recognition}, 5238--5248.

\bibitem[{Tong et~al.(2024{\natexlab{a}})Tong, Brown, Wu, Woo, Middepogu, Akula, Yang, Yang, Iyer, Pan et~al.}]{tong2024cambrian}
Tong, S.; Brown, E.; Wu, P.; Woo, S.; Middepogu, M.; Akula, S.~C.; Yang, J.; Yang, S.; Iyer, A.; Pan, X.; et~al. 2024{\natexlab{a}}.
\newblock Cambrian-1: A fully open, vision-centric exploration of multimodal llms.
\newblock \emph{arXiv preprint arXiv:2406.16860}.

\bibitem[{Tong et~al.(2024{\natexlab{b}})Tong, Liu, Zhai, Ma, LeCun, and Xie}]{tong2024eyes}
Tong, S.; Liu, Z.; Zhai, Y.; Ma, Y.; LeCun, Y.; and Xie, S. 2024{\natexlab{b}}.
\newblock Eyes wide shut? exploring the visual shortcomings of multimodal llms.
\newblock In \emph{Proceedings of the IEEE/CVF Conference on Computer Vision and Pattern Recognition}, 9568--9578.

\bibitem[{Wang et~al.(2024)Wang, Pan, Shi, Lu, Ren, Zhou, others, and Li}]{wang2024measuring}
Wang, K.; Pan, J.; Shi, W.; Lu, Z.; Ren, H.; Zhou, A.; others; and Li, H. 2024.
\newblock Measuring Multimodal Mathematical Reasoning with MATH-Vision Dataset.
\newblock \emph{Advances in Neural Information Processing Systems}, 37: 95095--95169.

\bibitem[{Weyand et~al.(2020)Weyand, Araujo, Cao, and Sim}]{weyand2020google}
Weyand, T.; Araujo, A.; Cao, B.; and Sim, J. 2020.
\newblock Google landmarks dataset v2-a large-scale benchmark for instance-level recognition and retrieval.
\newblock In \emph{Proceedings of the IEEE/CVF conference on computer vision and pattern recognition}, 2575--2584.

\bibitem[{Wu et~al.(2023{\natexlab{a}})Wu, Yao, Zhang, Song, Ouyang, and Wang}]{wu2023gpt4vis}
Wu, W.; Yao, H.; Zhang, M.; Song, Y.; Ouyang, W.; and Wang, J. 2023{\natexlab{a}}.
\newblock GPT4Vis: what can GPT-4 do for zero-shot visual recognition?
\newblock \emph{arXiv preprint arXiv:2311.15732}.

\bibitem[{Wu et~al.(2023{\natexlab{b}})Wu, Zhang, Xiong, Oguz, Gee, and Nie}]{wu2023role}
Wu, Y.; Zhang, P.; Xiong, W.; Oguz, B.; Gee, J.~C.; and Nie, Y. 2023{\natexlab{b}}.
\newblock The Role of Chain-of-Thought in Complex Vision-Language Reasoning Task.
\newblock \emph{arXiv preprint arXiv:2311.09193}.
\newblock ArXiv:2311.09193 [cs.CV].

\bibitem[{Yao et~al.(2024)Yao, Huang, Wu, Zhang, Wang, Liu, Wang, Song, Feng, Shen et~al.}]{yao2024mulberry}
Yao, H.; Huang, J.; Wu, W.; Zhang, J.; Wang, Y.; Liu, S.; Wang, Y.; Song, Y.; Feng, H.; Shen, L.; et~al. 2024.
\newblock Mulberry: Empowering mllm with o1-like reasoning and reflection via collective monte carlo tree search.
\newblock \emph{arXiv preprint arXiv:2412.18319}.

\bibitem[{Yu et~al.(2025)Yu, Wei, Peng, and Belongie}]{yu2025benchmarking}
Yu, H.-T.; Wei, X.-S.; Peng, Y.; and Belongie, S. 2025.
\newblock Benchmarking Large Vision-Language Models on Fine-Grained Image Tasks: A Comprehensive Evaluation.
\newblock \emph{arXiv preprint arXiv:2504.14988}.

\bibitem[{Zhang et~al.(2024)Zhang, Unell, Wang, Ghosh, Su, Schmidt, and Yeung-Levy}]{zhang2024visually}
Zhang, Y.; Unell, A.; Wang, X.; Ghosh, D.; Su, Y.; Schmidt, L.; and Yeung-Levy, S. 2024.
\newblock Why are Visually-Grounded Language Models Bad at Image Classification?
\newblock \emph{arXiv preprint arXiv:2405.18415}.

\end{thebibliography}


\end{document}